\documentclass[11pt]{article}
\usepackage[T1]{fontenc}
\usepackage{lmodern}
\usepackage[margin=1in]{geometry}
\usepackage{microtype}
\usepackage{amsmath,amssymb}
\usepackage{graphicx,booktabs,array}
\usepackage[round,authoryear]{natbib}
\usepackage{caption}
\usepackage[colorlinks=true,allcolors=blue]{hyperref}
\graphicspath{{figures/}}

\title{Consistency Is Not Correction:\\Separating Confidence, Abstention, and Recovery in LLM Agents}
\author{Khushal Sethi}
\date{October 2026}
\begin{document}
\maketitle
\begin{abstract}
A controller that resamples an answer, escalates to a larger model, or abstains
is usually justified by a confidence signal. We argue that three claims about such
a signal rhyme but do not follow from one another --- it ranks answers
(\emph{discrimination}), it supports withholding some of them
(\emph{selective prediction}), it changes an answer for the better
(\emph{correction utility}) --- and we give an exact decomposition of the third
into gated recoveries minus gated harms, which yields a bound that fixes whether a
comparison can succeed before it is run. Because a gate that only replaces a
proposal can change an outcome only on a task the base policy already got wrong,
the number of initial errors caps the discordant pairs any correct gate can
produce; when that cap falls below the count a paired test needs, the experiment
cannot return a positive result for a working mechanism, nor distinguish one from
a gate that never fired. We measure all three targets on two model scales and find
the bound binds. In a small-model suite a perturbation-agreement verifier reaches
within-policy AUROC 0.990 (95\% interval $[0.95, 1.00]$) while every episode
outcome matches single-sample decoding at 4.32 times the call count. In a
controlled 840-episode study no verifier replaced the proposal on any of the 60
GSM8K decisions, so that comparison tested nothing; the entire GSM8K gain across
all six controllers is one task, and it is the same task whose identical
temperature-zero repeats disagreed, making it the one answer the service does not
reproduce. On the diagnostic set the gate could reach four errors while the test
needed six, so even a perfect gate had no significant result available. We report
two checks that would have caught this before any budget was spent, and we show
that three routine analysis choices --- tie handling in rankings, tied selection
cutoffs, and unequal output limits --- each move a headline number in the same
dataset. The binding constraint on this literature is measurement resolution, not
the choice of signal.

\end{abstract}

\section{Introduction}
A language model that samples many answers and checks how much they agree can
report something single-sample decoding cannot: an estimate of its own
uncertainty. Systems increasingly act on that estimate, resampling an answer,
escalating to a larger model, or abstaining. The step from a score to an action
is treated as straightforward, and it is not. A confidence score can order
correct answers above incorrect ones while every replacement it triggers is also
incorrect, and abstaining can raise accuracy among the answers that are kept
without repairing a single one. These are different claims about a signal, they
are measured by different statistics, and we show they can disagree sharply on
the same data.

This paper makes that distinction precise and then measures it. We separate three
targets: \emph{discrimination}, whether a score ranks correct answers above
incorrect ones; \emph{selective prediction}, the accuracy retained as coverage is
traded away; and \emph{correction utility}, whether spending additional inference
actually changes task outcomes. The first two have well-developed statistics. The
third does not, so we give an exact decomposition of a controller's accuracy
change into gated recoveries minus gated harms that also fixes the cost at which
an intervention is worth taking. The decomposition is elementary, and that is the
point: it exposes an evaluation gap, not a modelling difficulty. A score with
perfect discrimination can have zero correction utility, and we observe exactly
that.

The decomposition does more than separate the targets. It bounds the experiment.
A gate that only ever replaces a proposal can change an outcome only on a task the
base policy already answered incorrectly, so the number of initial errors caps the
number of discordant pairs any correct gate can produce, and hence caps the effect
size the design can possibly register. Combined with the paired-evaluation
resolution diagnostic of \citet{kotawala2026resolution}, this turns into a
pre-registration check that costs nothing: count the errors, count the discordant
pairs the test needs, and if the first is smaller than the second the comparison
cannot succeed and should be redesigned rather than run. We were not clever enough
to apply our own check in advance, and the controlled suite we did run falls on
the wrong side of it.

To measure all three targets we instrument a family of inference-time
controllers, including prompt-perturbation verification, adaptive
self-consistency, an online signal-selection controller, and token
log-probability, and we report each one together with the rate at which its gate
actually fired. That last number is the second cheap check, and in our study it
is the decisive one: three verifier-driven policies never replaced the proposal
on any of the 60 GSM8K decisions, so that comparison was incapable of testing
their mechanisms at all.

Our contributions are:

\begin{itemize}
\item A three-target framework for inference-time reliability, including an exact
recoveries-minus-harms decomposition of correction utility with an explicit
cost condition (Section~\ref{sec:framework}).
\item An error-budget bound derived from that decomposition: initial errors cap
the discordant pairs a gate can produce, so a suite with too few errors cannot
return a significant result for a correct mechanism or distinguish it from an
inert one (Section~\ref{sec:framework}).
\item A resolution analysis that applies the bound to our own design, showing the
comparison required six discordant pairs while the error budget allowed four, so
no positive outcome was available (Section~\ref{sec:resolution}).
\item A decisiveness audit that records whether the mechanism under test ever
executed, and an attribution procedure that separates a controller's effect from
resampling alone and from base-call nondeterminism (Sections~\ref{sec:design}
and~\ref{sec:results}). In the controlled suite it shows the gate firing on none
of the GSM8K decisions and on $1.3$--$4.3\%$ of MiniHouse decisions, and shows
that every GSM8K outcome change traces to one answer the service does not
reproduce.
\item A demonstration that three routine analysis choices --- tie handling in
rankings, tied selection cutoffs, and unequal output limits --- each move a
headline number across a substantive threshold in the same dataset, with
tie-aware statistics and matched caps as the fix.
\end{itemize}

We report the results as they stand, including the ones that weaken a claim we
would have preferred to make. The signals evaluated here are not a general
indictment of consistency methods; they are a demonstration that the standard way
of evaluating them cannot distinguish a working mechanism from an absent one, and
that the standard way of reporting them does not say so.

\section{Related Work}
Self-consistency samples multiple reasoning paths and aggregates their
answers~\citep{wang2023selfconsistency}, and Adaptive-Consistency reduces its cost
by stopping when sampled answers agree~\citep{aggarwal2023adaptive}.
Adaptive action agreement extends the same idea to agent decisions, and we use it
as one of the evaluated policies~\citep{sethi2026trace}. Early stopping and
plurality voting are prior art; our interest here is what the agreement score can
be used \emph{for} once it is computed.

Two further observables are close to ours and distinct from them. SelfCheckGPT
uses disagreement between sampled generations to detect hallucination without
reference text~\citep{manakul2023selfcheck}, and semantic uncertainty aggregates
samples that mean the same thing rather than samples that are literally
equal~\citep{kuhn2023semantic}. Our numeric-answer experiments use a deterministic
parser instead of a learned equivalence model, so our estimates are coarser than
semantic uncertainty and we do not identify perturbation agreement with semantic
entropy. Prompt wording alone changes model outputs and measured accuracy
\citep{sclar2024quantifying}, which is why our perturbation probes retain the
exact prompts they issue rather than treating rewording as neutral. Asking a
model to evaluate its own proposed answer is a related but distinct observable
from agreement between samples~\citep{kadavath2022language}.

Discrimination statistics come with the usual caveat that confidence is not
correctness: we use ``calibration'' only in its established technical
sense~\citep{guo2017calibration} and do not claim it for scores evaluated as
rankings. Selective classification trades coverage for retained-answer
accuracy and has a developed theory of risk--coverage
trade-offs~\citep{geifman2017selective}; we use its descriptive machinery, but our
retrospective curves only describe a ranking. We fit no deployment threshold, hold
out no labels for one, and therefore claim no distribution-free correctness
guarantee.

\paragraph{Agreement is not a reliable confidence signal.}
We are not the first to report that agreement fails to track correctness in
agentic settings, and we do not claim that finding as novel. \citet{mehta2026disagree}
show that behavioral variance across runs is a usable selective-classification
signal while plurality voting adds only $0$--$2$ points in multi-step agents,
because early errors are committed rather than sampled away; their measurement of
trajectory divergence is closer to a diagnostic than ours. \citet{bahuguna2026backfires}
find that voting reduces per-problem accuracy on most hard science problems for
small models and that a verifier-free agreement gate captures almost none of the
headroom. \citet{ding2026agree} audit agreement across runners and prompts and
conclude it is a regime-dependent proxy, weakest for over-confident frontier
models. Our results are consistent with all three. What we add is not another
negative finding but a reason the comparisons that produce such findings may have
had no alternative: the error-budget bound in Section~\ref{sec:framework} and the
resolution analysis in Section~\ref{sec:resolution}, together with the
intervention rate as a required diagnostic. Those are design and reporting
instruments rather than claims about any particular signal.

\paragraph{Power and resolution in paired evaluation.}
Paired designs with binary outcomes should be tested on discordant
pairs~\citep{dietterich1998approximate, dror2018hitchhiker}, and NLP comparisons
are routinely underpowered~\citep{card2020littlepower}. \citet{kotawala2026resolution}
formalize required sample size, minimum detectable effect and the resolution ratio
$q=N/N^{\star}$ for paired LLM evaluation, and show that many public leaderboard
pairs do not meet a conventional target. We adopt that diagnostic directly and
cite it as theirs. Our addition is specific to reliability controllers: the
recoveries-minus-harms decomposition does not merely motivate a power calculation,
it upper-bounds the discordant pairs such a controller can generate, so the error
count of the base policy is itself the binding constraint. Generic resolution
analysis cannot see this, because it does not know which pairs a treatment is
capable of changing.

\section{Framework}
\label{sec:framework}
\subsection{Three Separate Questions}
Let $a_0(x)$ be an initial answer to task $x$, $Y_0(x)\in\{0,1\}$ its correctness,
and $s(x)$ a confidence score with larger values indicating greater confidence.
\emph{Discrimination} asks whether $s$ ranks correct answers above incorrect
answers. We report tie-aware AUROC,
\begin{equation}
 \Pr(s^+>s^-)+\tfrac12\Pr(s^+=s^-),
\end{equation}
when both outcome classes are present. A single-class sample has undefined
AUROC, rather than perfect AUROC. These ranking statistics do not establish
probability calibration.

\emph{Selective prediction} retains answers in a set $A$ chosen using their
scores. Its coverage is $\Pr(x\in A)$ and its risk is
$\Pr(Y_0=0\mid x\in A)$. At an exact retrospective coverage, a cutoff can split
a tied score group. We report the expected accuracy under label-independent
random tie resolution, together with the attainable minimum and maximum over
tie orders. The discrete area under the risk--coverage curve averages expected
prefix risk over all nonempty retained prefixes.

\emph{Correction utility} asks whether spending additional inference changes
task outcomes. For a one-step intervention, let $g(x)\in\{0,1\}$ indicate that
a gate requests a replacement $a_1(x)$, with correctness $Y_1(x)$. Its accuracy
change satisfies the identity
\begin{align}
 \Delta &= \mathbb{E}[g(Y_1-Y_0)]\\
 &= \Pr(g=1,Y_0=0,Y_1=1)
   -\Pr(g=1,Y_0=1,Y_1=0).
\label{eq:decomposition}
\end{align}
The first term counts gated recoveries and the second counts gated harms.
Good error detection is insufficient when replacements rarely recover errors,
or when changes to correct answers cause offsetting harm. For sequential agents,
different policies can visit different states; we therefore evaluate episode
outcomes directly and do not interpret unmatched trajectories as paired
state-level causal evidence.

\paragraph{A decision criterion, not a new theorem.}
Write $e=\Pr(Y_0=0\mid g=1)$, $r=\Pr(Y_1=1\mid Y_0=0,g=1)$,
and $h=\Pr(Y_1=0\mid Y_0=1,g=1)$. Then
\begin{equation}
 \Delta=\Pr(g=1)\{er-(1-e)h\}.
\end{equation}
If a correct result has utility one and each triggered intervention has
expected cost $c$ in the same utility units, intervening is beneficial only
when $er-(1-e)h>c$. AUROC alone determines none of $r$, $h$ or $c$.
For example, a score that perfectly flags initial errors has AUROC one,
but replacing each flagged error with another error gives $r=0$ and
$\Delta=0$ at positive cost. This elementary counterexample motivates the
measurement design; the decomposition is not claimed as a novel estimator
or a guarantee about any particular model.

\paragraph{The error budget.}
The decomposition yields a bound that constrains experimental design before
any data are collected. A gate that leaves correct answers untouched can
change an outcome only on a task the base policy already answered incorrectly,
so on a suite of $n$ tasks with $m$ initial errors the number of gated
recoveries is at most $m$, and the total number of discordant pairs available to
any such gate is at most $m$. Consequently a treatment that only ever replaces
a proposal can produce at most $m$ one-directional changes, and a paired test
that requires $d^\star$ discordant pairs to reject the null is
\emph{decided in advance} whenever $m<d^\star$: no correct gate can return a
significant result, and no correct gate can be distinguished from an inert one.
The same argument bounds selective prediction, because a gate that withholds
answers rather than replacing them cannot raise the number of correct answers at
all. This is why we report the initial error count as a first-class quantity
rather than a nuisance, and why Section~\ref{sec:resolution} evaluates designs
against it. The bound is a statement about the design, not about the mechanism:
it says nothing about whether the signal is good, only that the comparison was
never able to find out.

\subsection{Policy Families and Signals}
\label{sec:policies}
We instantiate the three targets with five policy families and five confidence
signals. Each family is a concrete, reproducible procedure; the labels below name
the mechanism, and a mechanism that passes one target need not pass another.

\paragraph{Baselines.} Greedy decoding makes one proposal per decision. Standard
or adaptive self-consistency draws $k$ stochastic proposals and returns their
plurality, with TrACE-style agreement stopping after two samples and expanding to
four while modal agreement stays below $0.75$~\citep{sethi2026trace}. We include
both $k{=}4$ and $k{=}5$ so that a verifier's real cost is bracketed by voting at
its typical and its maximum call allowance.

\paragraph{Prompt-perturbation verification (SPECTRA).} The verifier makes a
greedy proposal, queries three rewordings of the prompt at temperature zero, and
scores the fraction that reproduce the proposal. If the score falls below $0.5$ it
replaces the proposal with a stochastic sample. The point of the rewording is to
hold the task fixed and vary only the surface form, so we implement two variants
that differ in what they hold fixed. The aggressive variant also deletes recent
history, substitutes synonyms and pads the prompt with a restatement; these
operations are not assumed to preserve task-relevant information. The
conservative variant changes only output-instruction wording or prepends a
problem-solving instruction, leaving task facts and history intact. Distinguishing
these two is how we test whether an earlier negative outcome was an artifact of
the perturbation operator rather than of verification itself.

\paragraph{Online signal selection (COMPASS).} A meta-controller selects between
sampling agreement and perturbation agreement per decision using Beta-posterior
Thompson sampling, updating the posterior within an episode from whether the
committed action was legal, and resetting it at episode boundaries. Legality is
not task progress, and when the selected score clears the commit threshold the
controller keeps the greedy proposal, so agreement with an alternative modal
action does not force that action. Its arm-selection RNG is seeded independently
per task and repetition, and on one-step GSM8K it cannot observe feedback before
its only decision. We evaluate this implementation as written; its behaviour does
not bound what signal composition could achieve in general.

\paragraph{Post-hoc selection and free signals.} Confidence-gated selection keeps
the answers whose mean decision agreement over an episode exceeds a cutoff. Mean
output-token log-probability is read from the same greedy generation wherever the
endpoint exposes it, so it costs no additional generation call, though we claim no
zero implementation or provider overhead. Missing values stay missing. This
statistic is neither token entropy nor a calibrated probability that an answer is
correct, and we use it only as a ranking.

\section{Experimental Design}
\label{sec:design}
\paragraph{Benchmarks.}
MiniHouse is a deterministic text environment with two house layouts and discrete
navigation, manipulation and examination actions; task success is the
environment's goal predicate. It is not ALFWorld and supports no claim about
embodied robotics. GSM8K supplies arithmetic word problems with numeric gold
answers~\citep{cobbe2021gsm8k}. Both benchmarks are small and synthetic by design:
they are cheap enough that we can measure the triggers and costs of many policies
rather than the accuracy of one.

\paragraph{Parsing.}
A single deterministic parser scores every condition. For GSM8K it searches the
last recognized answer marker and otherwise the last numeric token, removes comma
grouping, and compares against gold with an absolute tolerance of $10^{-4}$; an
unparseable response is neither correct nor evidence of agreement with any
candidate. For MiniHouse an episode succeeds exactly when the environment's goal
predicate holds after replaying the parsed actions. Because the parser is shared,
a parsing bias affects all policies equally, but a malformed numeric response that
the parser misreads can still enter a paired comparison; we report parse failures
rather than dropping them.

\paragraph{Two suites.}
We report two suites that differ in model, scale and role. The \emph{small-model
suite} runs Qwen2.5-3B across 30 MiniHouse task IDs (three repetitions for
verification, one for the meta-controller) and 50 GSM8K problems, and serves to
establish the discrimination/utility gap and the effect of the analysis choices in
Section~\ref{sec:audit}. Deduplicating task content reveals that those 30 IDs
represent only 26 distinct layout--object--goal configurations, so all clustered
statistics use configurations, not IDs, as the unit of resampling.

The \emph{large-model suite} is the controlled comparison, and its protocol was
frozen before any of its results existed. It uses 60 GSM8K problems and 60 unique
MiniHouse configurations, one repetition, and seven policies per task (840
episodes). GSM8K tasks exclude every problem used in the small-model suite.
MiniHouse partitions hold 12 pilot, 11 unused calibration and 80 main
configurations, partitioned by a fixed shuffle; we take the first 60 main
configurations, disjoint from the pilot, though both draw on the same two
templates. All 103 enumerated placement configurations admit a constructive
solution of at most eight actions against an episode cap of 15, verified by a
scripted oracle with full task-state knowledge. That check establishes solvability,
not benchmark realism: a shortest path of eight actions still leaves little room
for a parsing or navigation error before the cap, so MiniHouse success rates should
be read as a property of this cap rather than of the task family.

The controlled suite uses \texttt{deepseek/deepseek-v4.1-flash} pinned to
Cloudflare through OpenRouter~\citep{openrouter2026api}. The identifier names an
API route, not an independently verified weight revision. Dedicated reasoning is
disabled while ordinary textual reasoning stays permitted; the completion cap is
1,024 tokens for GSM8K and 128 for MiniHouse in every condition, so output limits
cannot differ across policies. Stochastic proposals use temperature $0.7$ and
greedy, repeat and perturbation probes use temperature zero; seed parameters are
sent where the endpoint supports them, and API nondeterminism remains possible.
Jobs are shuffled under a fixed seed. Complete prompts, outputs, finish reasons,
logical calls, returned usage, provider identity and timing are retained, and no
provider fallback is permitted. A transport pilot over three model families
(168 episodes) is reported only as exploratory, and pilot tasks are excluded from
every controlled estimate.

\paragraph{Common-candidate diagnostics.}
For each main GSM8K greedy answer, we separately collect three stochastic
answers, three controlled prompt variants, three identical-input greedy
repeats, and one independent stochastic replacement. All signals target the
\emph{same} greedy answer. Agreement is the fraction of parsed answers equal
to that candidate; unparseable answers do not support it. We also compute
negative categorical entropy of the three stochastic numeric answers and mean
returned token log-probability of the original answer, when available.
The independent replacement permits a paired, one-step gating decomposition
at the fixed agreement threshold $0.5$. It is a diagnostic intervention, not
a claim that its seed or output reproduces an online perturbation-verification
replacement. The threshold $0.5$ is the only cutoff used for the agreement gates
and is fixed before analysis rather than tuned, because with three probes a support
value can only take the four levels $0$, $1/3$, $2/3$ and $1$, and $0.5$ is the
natural boundary between minority and majority reproduction. This coarseness is
itself a limitation of the diagnostic and is visible in the results.

\paragraph{Decisiveness.}
A controller that shares the base proposal but never fires its verifier cannot
change any outcome, so paired success differences cannot test its mechanism. We
therefore record, per arm, whether any step replaced the proposal
(\texttt{resampled}), and we report the fraction of decisions on which each arm
intervened alongside its outcome table. We also record, for every arm, whether its
first action coincides with the cached greedy action for the same task, because
each arm issues its own base call and the service does not guarantee that
identical temperature-zero requests return identical answers. Paired counts are
correspondingly reported as one-sample differences from the cached baseline
rather than as a two-sample contrast between policies.

\paragraph{Analysis.}
We report pooled episode success and mean logical calls alongside actual returned
tokens and dollars, where a logical call is one requested completion rather than
one HTTP batch or retry. Policy differences are paired by task and repetition.
Bootstrap intervals use 4,000 task-cluster resamples under a fixed analysis seed;
clusters merge duplicate configurations so that repeated content is not counted
as independent evidence. The diagnostic AUROC bootstrap uses 2,000 task resamples
and skips replicates with a single outcome class, reporting their valid fraction.
Because the controlled suite has one repetition, every intervention effect here
is estimated at $n{=}1$ per task and should be read as a plausible range rather
than a calibrated estimate.

These intervals are descriptive: we apply no multiplicity correction across the
twelve policy--benchmark comparisons, and we do not treat them as draws from a
population model for a finite synthetic benchmark. A bootstrap that collapses
after zero discordant outcomes does not establish equivalence, so where that
happens we report $1-0.05^{1/n}$ as a one-sided 95\% upper bound on cluster
discordance, which also bounds the absolute mean effect. We fit no deployment
threshold and claim no prospective selective-risk guarantee.

\section{Resolution of a Paired Reliability Comparison}
\label{sec:resolution}
The comparisons in this paper are paired at the task level and the outcome is
binary, so the relevant test is on discordant pairs rather than on two
independent accuracy estimates. This is standard paired-evaluation practice
\citep{dietterich1998approximate, dror2018hitchhiker}, and the reporting
protocol we use is the resolution diagnostic of \citet{kotawala2026resolution},
who give the required sample size $N^{\star}$ and the resolution ratio
$q=N/N^{\star}$ for exactly this setting. We adopt their diagnostic rather than
propose a new one; what is new here is applying it to inference-time reliability
controllers, where the error budget of Section~\ref{sec:framework} supplies a
design-specific cap that a generic resolution analysis does not know about.

\subsection{What the Design Could Detect}
A paired exact test at $\alpha=0.05$ needs
$6$

one-directional discordant pairs to reject the null. With 60 tasks per
benchmark, that is a minimum detectable effect of $10$ percentage points. The
controlled suite's power against a genuine improvement is therefore

\begin{table}[htbp]
\centering\small
\caption{Power of the 60-task paired comparison against a true improvement of the stated size, and the number of tasks needed for 80\% power at that size. A test on discordant pairs requires at least 6 one-directional changes before it can reject at $\alpha=0.05$, so effects below the corresponding threshold are not merely underpowered but unreachable.}\label{tab:power}
\begin{tabular}{rrr}\toprule
Effect (pp) & Power at $n=60$ & Tasks for 80\% power \\
\midrule
1 & 0.000 & 800 \\
2 & 0.001 & 400 \\
3 & 0.009 & 280 \\
5 & 0.079 & 157 \\
10 & 0.563 & 78 \\
15 & 0.903 & 52 \\
\bottomrule\end{tabular}
\end{table}

A controller that truly improved $5\%$ of tasks would have been detected about
7.9\%

of the time; detecting such an effect reliably would have required
157

tasks per benchmark. The suite was sized for effects of roughly $15$ percentage
points and up, which is far larger than the differences this literature
typically reports and larger than anything the small-model suite shows.

\begin{figure}[htbp]
\centering\includegraphics[width=0.98\linewidth]{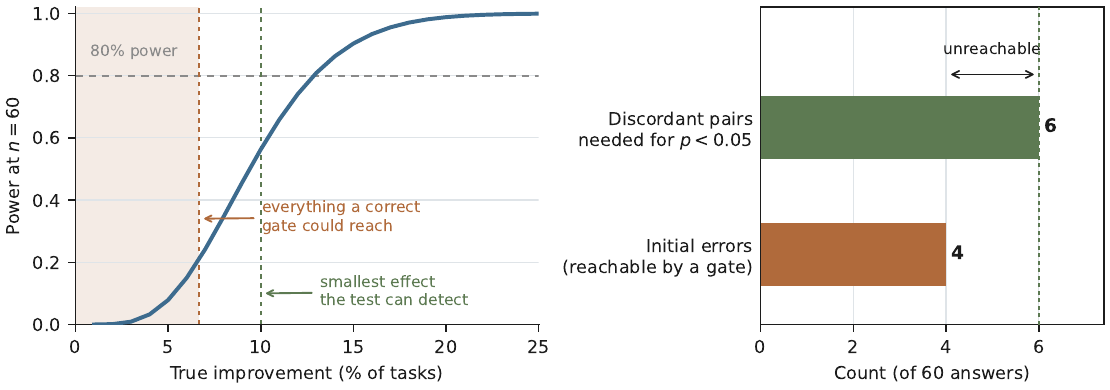}
\caption{Left: power of the 60-task paired test against a true improvement, with
the region a correct gate could reach shaded. The test needs six one-directional
changes to reject at $\alpha=0.05$, so it cannot detect anything below a ten-point
effect, and the error budget caps the reachable effect below that threshold. Right:
the two quantities that decide the outcome of the experiment, before any data are
collected.}
\label{fig:resolution}
\end{figure}

\subsection{The Error Budget Bounds the Reachable Effect}
The generic resolution calculation assumes an arbitrary effect size is
available. In fact it is not. Applying the bound from Section~\ref{sec:framework}
to the diagnostic set gives an exact ceiling:

\begin{table}[htbp]
\centering\small
\caption{The diagnostic error budget. A gate that only replaces an answer can change an outcome only where the initial answer was wrong, so the initial error count caps the number of gated recoveries and therefore the number of discordant pairs any such gate can produce.}\label{tab:budget}
\begin{tabular}{lr}\toprule
Quantity & Value \\\midrule
Greedy GSM8K answers scored & 60 \\
Incorrect answers (reachable by a gate) & 4 \\
Maximum gated recoveries & 4 \\
Discordant pairs required for $p<0.05$ & 6 \\
\bottomrule\end{tabular}
\end{table}

The gate can only replace an incorrect answer, so the diagnostic offers at most
four opportunities for a beneficial change across all 60 answers, while
significance requires six. Even a gate that recovered \emph{every} incorrect
answer it could reach would return
$p=0.125$

which does not clear the threshold. The comparison was therefore incapable of
producing a positive result for a correct mechanism, and equally incapable of
distinguishing such a mechanism from one that never fired. That is a stronger
statement than a null result: it is a statement that the experiment had no
positive outcome available to it.

\subsection{The Two Recommendations}
Read together, the framework and the resolution analysis give a checklist that
is cheap to apply before any generation budget is spent.

\begin{enumerate}
\item \textbf{Publish the error budget and the resolution ratio.} Report the
number of initial errors $m$ next to the number of tasks $n$, compute the
required discordant pairs, and state the design's power at the effect size of
interest. If $m$ is smaller than the required discordant count, the comparison
cannot succeed and should be redesigned rather than run. On our diagnostic set
the resolution ratio is
$q=60/157\approx0.38$ at a five-point effect

\item \textbf{Publish the intervention rate.} Report the fraction of decisions
on which the gate actually fired. A signal that never triggers is not evidence
about that signal, and a paired table without an intervention rate cannot
distinguish a silent gate from an ineffective one. This is the cheapest of all
the diagnostics and, in our suite, the decisive one: it is what revealed that
the GSM8K column tested nothing at all.
\end{enumerate}

Neither recommendation requires new inference, and either alone would have
prevented the central ambiguity of this study.

\section{Results}
\label{sec:results}
\subsection{Discrimination Without Correction}
\label{sec:audit}
\begin{figure}[htbp]
\centering\includegraphics[width=0.98\linewidth]{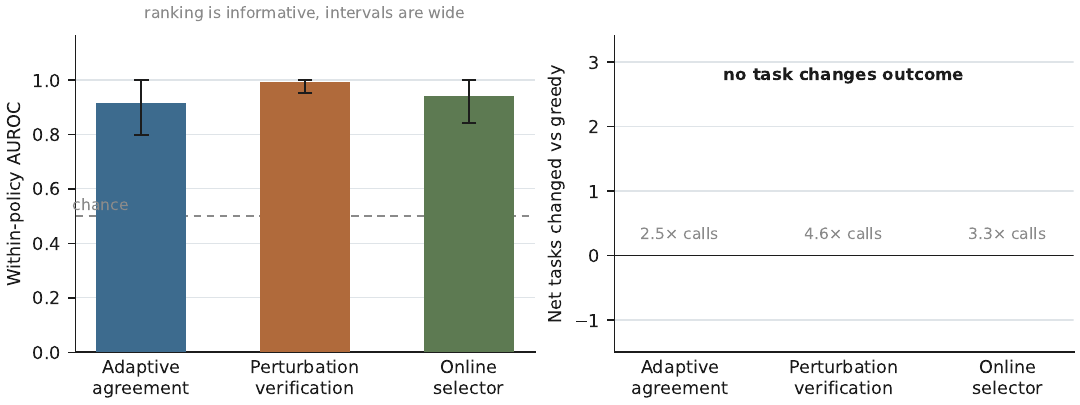}
\caption{Small-model MiniHouse experiment ($n=30$ task IDs, 26 distinct
configurations). Left: within-policy AUROC with task-cluster bootstrap
intervals; the ranking is informative but the intervals are wide and overlap.
Right: net tasks whose outcome changed relative to greedy, with realized call
ratios. All three policies rank well and none changes a single outcome while
spending 2.5--4.6 times the calls. AUROC measures discrimination, not correction
utility.}
\label{fig:separation}
\end{figure}

The clearest demonstration that the three targets come apart is the one-repetition
small-model MiniHouse experiment (Figure~\ref{fig:separation}). All four policies
succeed on exactly the same 11 of 30 task IDs, so no policy recovers or harms a
single task relative to greedy. Their episode-level agreement scores nonetheless
rank well against outcome: AUROC is 0.914 for adaptive agreement, 0.990 for
prompt-perturbation verification and 0.943 for the online selector, with
task-cluster bootstrap intervals of [0.80, 1.00], [0.96, 1.00] and [0.84, 1.00].
The ranking is real, but 11 successes among 30 episodes leave the intervals wide,
and the ordering between the top two policies is not resolved. Cost diverges just
as sharply, from 5.87 calls per episode for greedy to 14.67, 26.93 and 19.27, so
the strongest ranker is the most expensive and the cheapest policy is not the
weakest ranker. High discrimination and the complete absence of observed
correction coexist in the same data. These are within-policy rankings rather than
estimators applied to one shared answer, so they show that ranking information is
present while correction is not; they do not show that the scores \emph{cause} the
absence of correction.

The three-repetition verification experiment agrees. Greedy and
prompt-perturbation verification each succeed on 33 of 90 episodes, and their
outcomes match on every paired task--repetition, giving zero recoveries and zero
harms. Verification spends 25.34 calls against greedy's 5.87, a 4.32-fold
increase, and its higher-effort setting spends 30.46 calls for the same success
vector. Because those 90 episodes cover only 26 distinct configurations, zero
discordant clusters yields a conservative 95\% upper bound of 10.9 percentage
points on the population mean effect under the stated cluster assumption; the
zero-width percentile interval a naive bootstrap would produce instead would hide
that uncertainty entirely.

\subsection{Three Analysis Choices That Change the Conclusion}
Three ordinary choices in the analysis pipeline each move a headline number, and
two of them move it across a substantive threshold.

\emph{Tie handling.} On the 50-problem GSM8K self-consistency run, correcting
AUROC for tied scores raises the four-sample score from 0.870 to 0.919 and the
eight-sample score from 0.926 to 0.945. The correction strengthens rather than
weakens discrimination, which is the useful outcome: the defect was
conservative, and a paper reporting the uncorrected numbers would understate its
own signal. It does not, however, validate a correction policy.

\emph{Tied cutoffs.} Four-sample selection retains 30 of 50 answers for 100\%
accuracy in stored order, but the cutoff falls inside an eight-answer tie. Under
label-independent tie resolution the accuracy is 99.17\%, and the attainable
range over tie orders is 96.7--100\%, so a reported 100\% is an artifact of
stored order. The corresponding 60\%-coverage result at eight samples is robust
to its own boundary tie in this sample, though it remains a finite-sample
observation rather than a guarantee.

\emph{Unit of analysis and output limits.} The larger saved file holds 300 greedy
rows but only 236 four-sample rows, so it is not a complete policy matrix, and
scoring its incomplete sample against all 300 greedy rows would silently change
the task distribution. Reported on its own 236 tasks, four-sample selection
succeeds on 83.5\% with a tie-corrected AUROC of 0.862; at its stored 142/236
coverage the expected accuracy under random ties is 97.27\%, with a tie-order
range of 95.07--97.89\%. Separately, mean token log-probability attains AUROC
0.720 over 150 records from 50 tasks at 84\% accuracy, but that run allowed 512
output tokens while its comparison baseline allowed 384, so its gap from the
baseline cannot be attributed to observing log-probabilities, and only output
prefixes were stored, so the completions cannot be regraded. As a final
consistency check we replay 2,008 saved records and find no disagreement between
stored labels and stored actions; this verifies the pipeline, not the provenance
of the original model calls.

\subsection{Controlled Comparison}
\label{sec:controlled}
Every policy in the controlled suite issues its own base call, and the service
does not guarantee that identical temperature-zero requests return the same answer
(Section~\ref{sec:diag}), so the paired table below is a bounded replication of
the small-model pattern rather than a clean between-policy experiment. The
decisiveness audit that follows states what the arms actually did.
\begin{table}[htbp]
\centering\small
\caption{GSM8K: 60 tasks per policy, one repetition. Calls and tokens are means per episode; cost is the total for each policy. Gain/harm counts paired outcome changes relative to greedy. Tokens include prompt and completion usage.}\label{tab:gsm8k}
\begin{tabular}{lrrrrr}\toprule
Policy & Success (\%) & Calls & Tokens & Cost (\$) & Gain/harm \\
\midrule
Greedy & 93.3 & 1.00 & 257 & 0.0090 & --- \\
SC-4 & 93.3 & 4.00 & 1045 & 0.0372 & 1/1 \\
SC-5 & 95.0 & 5.00 & 1304 & 0.0463 & 1/0 \\
TrACE-4 & 95.0 & 2.00 & 516 & 0.0182 & 1/0 \\
Perturb verify (aggressive) & 95.0 & 4.00 & 1057 & 0.0372 & 1/0 \\
Perturb verify (conservative) & 95.0 & 4.00 & 1058 & 0.0368 & 1/0 \\
Online selector & 95.0 & 3.00 & 778 & 0.0273 & 1/0 \\
\bottomrule\end{tabular}
\end{table}
\begin{table}[htbp]
\centering\small
\caption{MiniHouse: 60 tasks per policy, one repetition. Calls and tokens are means per episode; cost is the total for each policy. Gain/harm counts paired outcome changes relative to greedy. Tokens include prompt and completion usage.}\label{tab:minihouse}
\begin{tabular}{lrrrrr}\toprule
Policy & Success (\%) & Calls & Tokens & Cost (\$) & Gain/harm \\
\midrule
Greedy & 86.7 & 6.25 & 1129 & 0.0076 & --- \\
SC-4 & 91.7 & 25.33 & 4580 & 0.0304 & 3/0 \\
SC-5 & 88.3 & 32.58 & 5890 & 0.0388 & 1/0 \\
TrACE-4 & 93.3 & 13.93 & 2529 & 0.0166 & 4/0 \\
Perturb verify (aggressive) & 91.7 & 24.93 & 4573 & 0.0310 & 3/0 \\
Perturb verify (conservative) & 90.0 & 25.48 & 4689 & 0.0316 & 2/0 \\
Online selector & 91.7 & 18.78 & 3412 & 0.0230 & 3/0 \\
\bottomrule\end{tabular}
\end{table}

On GSM8K, greedy already succeeds on 93.3\% of episodes, so the full spread across the six controllers is 93.3 to 95.0\% --- one to two tasks. Every controller that moves an outcome moves it in the favourable direction and none harms a task that greedy solved. The strongest net effect is 1 task (SC-5, TrACE-4, Perturb verify (aggressive), Perturb verify (conservative), Online selector), costing 2.00 to 5.00 times the greedy call count. SC-4 is the exception, trading one task for another (1 gain, 1 harm, +0.0~pp), which is what a single resampled swap looks like. No comparison on this task family separates from zero before any multiplicity correction. A single repetition cannot separate a policy effect from decoding noise, so even that difference is a hypothesis for a larger replication rather than an established gain.

On MiniHouse, greedy already succeeds on 86.7\% of episodes, so the full spread across the six controllers is 88.3 to 93.3\% --- one to two tasks. Every controller that moves an outcome moves it in the favourable direction and none harms a task that greedy solved. The strongest net effect is 4 tasks (TrACE-4), costing 2.23 to 2.23 times the greedy call count. Only TrACE-4 ([+1.7, +13.3]~pp) separates from zero before any multiplicity correction. A single repetition cannot separate a policy effect from decoding noise, so even that difference is a hypothesis for a larger replication rather than an established gain.

Three of the arms are verifier-driven: they can replace the base action, and their records state whether they did. On GSM8K, where each of these policies makes exactly one decision per episode, the verifier replaced the proposal in 0 of 60 decisions (aggressive perturbation verifier), 0 of 60 (conservative perturbation verifier) and 0 of 60 (online selector). On MiniHouse the same verifiers intervened on 13, 4 and 8 of 60 episodes, at roughly 1.3 to 4.3 per cent of decisions. In the remaining decisions these arms commit to the proposal they would have made without the signal, so their success vectors cannot differ from the baseline except through those interventions or through nondeterminism in the base call. The measured intervention rates bound the room for any per-decision effect: had these arms intervened at a one-per-cent rate instead, the probability of observing no intervention at all over the 60 GSM8K decisions would be 0.547. A null paired result from these arms therefore indicates that the implemented gate almost never fires, not that replacements were tested and found unhelpful, and the mechanism's utility is correspondingly untested at this operating point.

A second limitation limits what the paired counts can attribute. The base proposal is a separate API call in each arm, and identical temperature-zero inputs were observed to return different answers in one of the 60 diagnostic tasks, so the arms' first actions are not guaranteed to coincide with the greedy baseline. The number of episodes whose first action differs from the cached greedy action is SC-4: 9 MiniHouse and 2 GSM8K, TrACE-4: 9 MiniHouse and 1 GSM8K, Perturb verify (conservative): 1 MiniHouse and 1 GSM8K. A paired comparison across separately issued calls therefore mixes the controller's effect with base-call nondeterminism, which is why we report paired counts as one-sample rather than two-sample statistics. The single GSM8K paired gain in Table~\ref{tab:gsm8k} comes from exactly the one GSM8K task whose identical-input repeats disagreed, so that difference should be attributed to generation noise rather than to any controller.

On MiniHouse the largest mean paired difference belongs to TrACE-4 (+6.7~pp), which never fires a verifier at all. The arms that never fire one (SC-4, SC-5, TrACE-4) recover 5 distinct tasks between them, and the verifier-driven arms (Perturb verify (aggressive), Perturb verify (conservative), Online selector) recover 3 tasks, of which 3 are in both sets. The verifier-driven arms therefore recover a subset of the tasks that resampling and voting already recover here. These arms differ from greedy only by drawing several proposals and voting, so the paired improvement in this column is available without any confidence signal, and the verifier's marginal contribution is not identified against a single-sample baseline.

One number in that table carries almost all of the apparent signal. On GSM8K the
union of paired gains across all six controllers is a single task
(\texttt{gsm8k\_test\_696}), and it is the same task whose identical-input
temperature-zero repeats disagreed in Section~\ref{sec:diag}. No other GSM8K
answer changes outcome under any policy, so the only policy-visible movement on
this benchmark is the one answer the service does not reproduce. The single harm,
in four-sample voting, is likewise one answer (\texttt{gsm8k\_test\_87}). Any
GSM8K improvement here should therefore be attributed to generation
nondeterminism rather than to a controller, which is why we do not report a
positive GSM8K result at all.

\subsection{How Often Each Gate Fired}
The paired tables above are only interpretable alongside the rate at which each
gate actually executed, because a gate that never fires cannot change an outcome.

\begin{figure}[htbp]
\centering\includegraphics[width=0.98\linewidth]{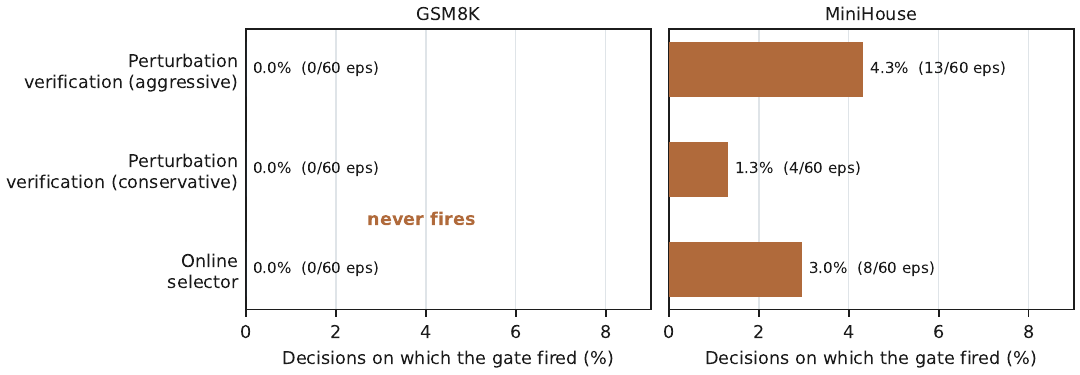}
\caption{Fraction of decisions on which each verifier-driven policy replaced the
base proposal, in the controlled suite. On GSM8K all three are exactly zero, so
that column tests an inert gate rather than a rejected mechanism. On MiniHouse the
rates are $1.3$--$4.3\%$, which bounds the room any per-decision effect had.}
\label{fig:intervention}
\end{figure}

\subsection{Signal Performance on a Common Candidate}
\label{sec:diag}
\begin{table}[htbp]
\centering\small
\caption{All signals score the same 60 greedy GSM8K answers. AURC averages expected risk over nonempty prefixes, with label-independent randomized ties. Smaller AURC is better.}\label{tab:diagnostics}
\begin{tabular}{lrrrr}\toprule
Signal & $n$ & AUROC & 95\% interval & AURC \\
\midrule
Stochastic support & 60 & 0.750 & [0.500, 1.000] & 0.035 \\
Perturbation support & 60 & 0.750 & [0.500, 1.000] & 0.035 \\
Identical-input repeat & 60 & 0.625 & [0.500, 1.000] & 0.051 \\
Mean token log-prob. & 60 & 0.821 & [0.475, 1.000] & 0.019 \\
Negative answer entropy & 60 & 0.625 & [0.500, 0.912] & 0.051 \\
\bottomrule\end{tabular}
\end{table}
\begin{table}[htbp]
\centering\small
\caption{Independent-replacement diagnostic at the fixed support threshold 0.5. Gains and harms refer to the same saved candidate and a separately sampled replacement, not matched sequential states.}\label{tab:gating}
\begin{tabular}{lrrrr}\toprule
Gate & Triggered & Recovered & Harmed & $\Delta$ (pp) \\
\midrule
Stochastic support & 2 & 1 & 0 & +1.7 \\
Perturbation support & 2 & 1 & 0 & +1.7 \\
Identical-input repeat & 1 & 1 & 0 & +1.7 \\
\bottomrule\end{tabular}
\end{table}

\begin{figure}[htbp]
\centering\includegraphics[width=0.98\linewidth]{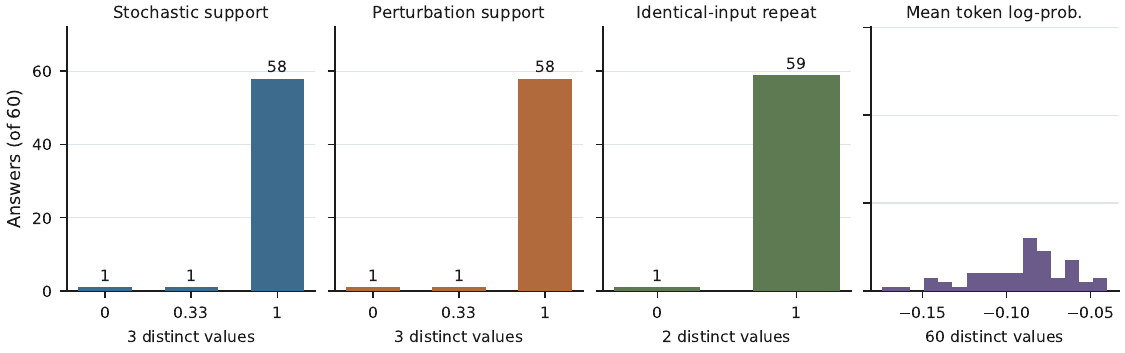}
\caption{Distribution of each confidence signal over the 60 common-candidate
GSM8K answers. Three of the four agreement signals take only two or three
distinct values on this sample, so their AUROC reflects how a large tied group is
resolved rather than a graded preference; only mean token log-probability varies
across all 60 answers.}
\label{fig:spread}
\end{figure}
All signals score the same 60 greedy GSM8K answers, of which only 4 are incorrect, so every AUROC rests on the same 4 errors and the ensemble of signals is describing one 60-answer sample rather than five independent measurement problems.

Stochastic support separates only 3 levels across those 60 answers (0, 0.33, 1), and is below 1 only on 2 tasks. It is near-constant here, so its AUROC of 0.750 rests on the same 2 tasks and reports the resolution of the tied group rather than a graded preference.

Perturbation support separates only 3 levels across those 60 answers (0, 0.33, 1), and is below 1 only on 2 tasks. It is near-constant here, so its AUROC of 0.750 rests on the same 2 tasks and reports the resolution of the tied group rather than a graded preference.

Identical-input repeat separates only 2 levels across those 60 answers (0, 1), and is below 1 only on 1 task. It is near-constant here, so its AUROC of 0.625 rests on the same 1 task and reports the resolution of the tied group rather than a graded preference.

A more specific warning applies to the two agreement probes. Resampling the same prompt at temperature 0.7 and resampling three reworded prompts at temperature zero produce the same reported support on all 60 answers even though they issue different requests. The stored records do not distinguish the two partitions, so on this sample the comparison between them carries no information; whether that is a genuine equivalence or a shared failure of the parser to separate them cannot be decided from the metadata we retained.

Mean token log-probability is the only signal with a distinct value on every answer, and it also has the highest AUROC of the set (0.821), but its interval [0.475, 1.000] still spans the range from chance.

At the fixed threshold 0.5 the three agreement gates trigger on 2 tasks, 2 tasks and 1, always within the same low-support group, and every gate recovers one answer without harming another. Across all three gates the triggered set has 2 distinct tasks, so the three gate rows are one event counted three ways.

With only 4 incorrect answers on this subset every bootstrap interval spans essentially the whole range, so the ranking gaps (best mean token log-prob. at 0.821, lowest identical-input repeat at 0.625) are not separable at this sample size, and neither are the gates. These are descriptions of what the mechanism does on four errors, not effect estimates.

The section still isolates the framework's distinction, but by construction rather than by graded evidence: a signal can rank the initial errors well and still trigger replacements whose outcomes match the initial answer, which is why discrimination, selective prediction and correction utility are separate targets. In the other direction the AUROC spread here is measured on one shared candidate and does not compare the policies themselves.

Because one task's candidate and its identical-input repeat disagree, that task is the only place where the initial answer is not stable, and it is therefore the only place the controlled comparison between policies can visibly move. We report that dependency explicitly rather than as a caveat: a reader should treat the controlled paired counts as describing the behaviour of one unstable answer and one stable remainder.

\begin{figure}[htbp]
\centering\includegraphics[width=0.62\linewidth]{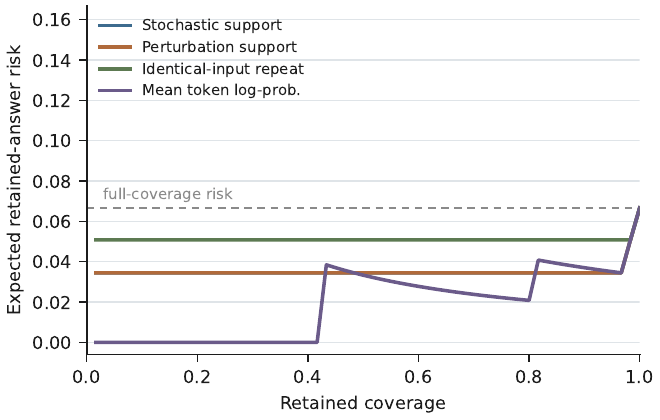}
\caption{Retrospective risk--coverage curves for the same main-run greedy GSM8K
answers. Tied scores are resolved in expectation without consulting labels. The
curves are nearly indistinguishable at low coverage because the signals disagree
on at most three answers. These are ranking descriptions, not deployable
thresholds chosen on unseen test labels.}
\label{fig:risk}
\end{figure}

\section{Discussion and Limitations}
\paragraph{Why detection and action must be evaluated together.}
The small-model result is not merely a nonsignificant improvement:
prompt-perturbation verification ranks episode outcomes well while preserving
exactly the same success/failure vector as greedy decoding.
Equation~\ref{eq:decomposition} explains the distinction without assuming that a
confidence score is useless. A deployment decision requires both a diagnostic and
a replacement that can resolve the detected error at an acceptable cost, and
retaining only high-score answers answers a different question. The evaluation
therefore has to report initial errors, triggered interventions, recoveries,
harms, and the inference actually spent, which is exactly the evidence a
discrimination-only report omits.

\paragraph{The gates were largely inert, and that is the main threat to
validity.}
The controlled suite was designed to remove implementation confounds, yet its
paired counts are far smaller than a mechanistic test should produce. The reason
is that the implemented gate almost never fires. None of the three verifier-driven
policies replaced the model's proposal on any GSM8K decision, and on MiniHouse they
fire on 1.3--4.3\% of decisions. On every decision where the gate does not fire,
the policy is the proposal it would otherwise have made, so no paired count can
reveal a mechanism that was never exercised. The negative reading of the table is
therefore unsupported: it shows an untested mechanism, not a tested and rejected
one. The small-model comparison has the same limitation compounded, because its
intervention rate was never recorded and its saved metadata keep only an inferred
resample flag, so the question cannot be settled retrospectively there either.

A sharper version of the same problem affects the headline numbers. Because the
service did not reproduce one GSM8K answer under identical temperature-zero
requests, that task is the only place where an initial answer is unstable, and it
is therefore the only place the controlled comparison can visibly move: all six
controllers gain on exactly that task and nowhere else. The paired counts on
GSM8K are consequently not a small sample of policy behaviour but a complete
description of generation nondeterminism, and we read them that way. This is worth
stating plainly because the same design would produce an apparently clean
positive result if the unstable task had happened to fall on the treatment arm
and the analysis stopped at the paired table.

The MiniHouse column also cannot be attributed to the signals. Arms that never
fire a verifier recover several of the same tasks as the verifier-driven arms, and
they differ from greedy only by sampling several proposals and voting. Most of the
paired improvement is therefore available from resampling and voting alone, which
is a competent baseline that a design centred on a confidence signal tends not to
isolate.

The one apparent correction has a simpler explanation than the controller. The
identical-input diagnostic shows that repeated temperature-zero requests to this
service do not always return the same answer, and the single GSM8K paired gain
falls on exactly that task. Paired counts computed across separately issued base
calls consequently mix the policy under test with generation nondeterminism, which
is why we report them as one-sample statistics rather than as a clean
between-policy contrast.

A capacity gap separates the two suites: 3B parameters locally against one large
API model, so they do not share a regime. Read together, the large-model runs fail
the resolution check of Section~\ref{sec:resolution} on both counts --- too few
errors to spend and too few decisions on which the gate can act --- while the
small-model runs show a mechanism whose implemented operator did not supply
correction. This is weaker than a demonstration that consistency signals fail; it
is a demonstration that the standard test of whether they fail cannot answer the
question at this operating point.

The stronger reading is the one the bound supports, and it is not limited to our
experiment. Any controller that shares the base proposal and selectively replaces
it can only change outcomes where the base policy was wrong, so its maximum
achievable effect is the base error rate; controllers that only abstain cannot
change outcomes at all. Reliability benchmarks are moving toward saturated
frontier models on which the base error rate falls well below the resolution
threshold of a paired test of conventional size, which means the field's default
comparison design will increasingly return nulls that carry no information about
the mechanisms under test. The remedy is not a larger budget spent on the same
design but a design chosen for its resolution: harder items, more of them, and a
reported intervention rate. What would change our conclusion is a suite with
hundreds of borderline items, sized against the bound rather than by convention,
and a comparison against matched-budget resampling and voting with no confidence
signal.

\paragraph{What this study does not identify.}
The policy families span more than one mechanism, so their outcome patterns
cannot isolate model capacity, signal dependence, prompt semantics, replacement
quality or the legality reward as a unique explanation. The controlled suite
removes several implementation confounds but uses one API model, one repetition
and two narrow task families, and its restricted perturbations do not exhaust
good paraphrases or verifier designs. The three-sample entropy estimate is coarse
and operates on parsed numeric classes rather than meaning. MiniHouse has only two
templates, so its 60 configurations are not 60 independently designed
environments. Seed control does not establish reproducible weights or
deterministic service execution, provider prices and behaviour can change, and
recorded dollars are specific to this experiment.

\paragraph{Limits of the small-model reanalysis.}
Replaying saved actions checks that stored labels are consistent with the stored
actions and environment; it does not authenticate the original generations.
Those files retain only a 200-character completion prefix, so full answer
extraction cannot be reconstructed, and a larger output cap in one run prevents
attributing an accuracy difference to log-probability access. Missing decision
metadata precludes reconstructing every gate. Confidence comparisons across
different policies target different answers and must not be read as a controlled
comparison of estimators; the common-candidate design addresses that target
mismatch for GSM8K only.

\paragraph{Practical implication.}
Three checks follow, in increasing order of cost, and all are cheaper than the
runs they would save. First, before deploying an additional inference mechanism,
record the decision it changes and compare it with a simple sampling baseline at
the relevant resource budget. Second, before running the comparison at all,
compute the error budget: if the base policy's error count on the suite is below
the discordant-pair count a paired test needs, no correct mechanism can be shown
to work and the suite must be replaced. Third, report the intervention rate
alongside the outcome table, because a gate that never fires turns every
downstream statistic into a description of the baseline. Report randomized-tie
selection curves rather than a favorable input order. When a small benchmark
saturates, report the resulting lack of information instead of treating undefined
AUROC or zero observed harms as proof of safety. These checks apply to evaluating
reliability mechanisms; our data do not establish reliability for silicon design,
medicine or other safety-critical applications.

\section{Conclusion}
Confidence ranking, selective prediction and correction utility are distinct
evaluation targets, and the decomposition of the third into gated recoveries
minus gated harms does more than separate them: it bounds the experiment. Because
a gate that only replaces a proposal can act only where the base policy was
already wrong, the initial error count caps the discordant pairs any correct gate
can produce, and therefore caps the effect a paired comparison can register. When
that cap falls below the number of discordant pairs a test requires, the
comparison cannot return a positive result for a working mechanism and cannot
distinguish one from a gate that never fired.

We measured all three targets across two model scales, and that bound is what our
own controlled study ran into. Three verifier-driven policies never replaced the
proposal on any of the 60 GSM8K decisions; the single GSM8K outcome change across
all six controllers traces to one answer that the service does not reproduce under
identical temperature-zero requests; and on the diagnostic set the gate could
reach four errors while the test needed six. The small-model suite supplies the
complementary evidence, with high discrimination, wide intervals, and no change
to any task outcome at a multiple of the call cost. Routine analysis choices
around tied rankings, tied cutoffs and unequal output limits each move a headline
number in the same data.

The conclusion we support is therefore about measurement as much as about signals.
The binding constraint on this literature is resolution, not the choice of
confidence signal: report the initial error budget against the discordant pairs
the design needs, report the rate at which the gate actually fires, treat the
replacement as carefully as the trigger, and compare against resampling and
voting at a matched budget before claiming that an uncertainty signal improves an
agent.

\paragraph{AI assistance.}
AI tools assisted with code, analysis and manuscript preparation. The submission
does not attribute authorship or independent experimental judgment to those tools.

\bibliographystyle{plainnat}
\bibliography{references}
\appendix
\section{Reproduction and Provenance}
The ancillary archive contains frozen benchmark partitions, the small-model
result snapshots with source hashes, analysis scripts, controller code, provider
metadata and the controlled result records. Credentials and local
credential-entry utilities are excluded. Recomputing the statistics requires
Python with NumPy and SciPy; rebuilding benchmark partitions additionally
requires the Hugging Face \texttt{datasets} package and network access, while the
frozen partitions suffice for every reported number. New inference needs an
independently configured account, and the service is not guaranteed to reproduce
the original outputs bit for bit, which is itself one of the findings.

\paragraph{Parsing and scoring.}
GSM8K parsing first searches the last recognized answer marker and then falls
back to the last numeric token; commas are removed and a $10^{-4}$ numerical
tolerance is used. This parser can misread malformed prose and is shared across
conditions. MiniHouse success is the environment's goal predicate after replay
of discrete parsed actions. Empty or invalid actions do not count as success.
These rules score task outputs, not reasoning faithfulness.

\paragraph{Controlled perturbations.}
GSM8K prepends one of: ``Solve the following problem.'',
``Work out the answer to this problem.'',
or ``Find the numerical solution to the following question.'' The exact
versioned strings and prompts are authoritative in the code and request logs.
MiniHouse changes only the response-format instruction, keeping observation,
goal and the five most recent actions. Such changes preserve the supplied facts
but may change attention, reasoning effort or adherence; semantic invariance of
the model's internal computation is not assumed.

\paragraph{Transport and missing data.}
Endpoint-specific capabilities determine whether seeds, reasoning controls and
log-probabilities are requested. Model-wide capability unions are insufficient.
Failed requests retain status and conservative charges; uncertain responses are
not silently retried as if free. Early endpoint probes and the incomplete
transport pilot are excluded from the controlled estimates. Main-run completeness, parse
failures and finish reasons are checked before packaging. Raw journals contain
provider responses rather than credentials. Reported sampling conditions are
logical policies; different services need not implement identical randomness.

\section{Additional Numerical Results}
\begin{table}[htbp]
\centering\small
\caption{Paired main-run task-bootstrap intervals versus greedy. These intervals are descriptive and are not corrected for multiple comparisons.}\label{tab:effects}
\begin{tabular}{lrrr}\toprule
Task & Policy & $\Delta$ (pp) & 95\% interval (pp) \\
\midrule
gsm8k & SC-4 & +0.0 & [-5.0, +5.0] \\
gsm8k & SC-5 & +1.7 & [+0.0, +5.0] \\
gsm8k & TrACE-4 & +1.7 & [+0.0, +5.0] \\
gsm8k & Perturb verify (aggressive) & +1.7 & [+0.0, +5.0] \\
gsm8k & Perturb verify (conservative) & +1.7 & [+0.0, +5.0] \\
gsm8k & Online selector & +1.7 & [+0.0, +5.0] \\
minihouse & SC-4 & +5.0 & [+0.0, +11.7] \\
minihouse & SC-5 & +1.7 & [+0.0, +5.0] \\
minihouse & TrACE-4 & +6.7 & [+1.7, +13.3] \\
minihouse & Perturb verify (aggressive) & +5.0 & [+0.0, +11.7] \\
minihouse & Perturb verify (conservative) & +3.3 & [+0.0, +8.3] \\
minihouse & Online selector & +5.0 & [+0.0, +11.7] \\
\bottomrule\end{tabular}
\end{table}
\begin{table}[htbp]
\centering\small
\caption{Transport pilot, aggregated across seven policies and only four tasks per benchmark. These are not independent trials or evidence of family superiority, and pilot tasks are excluded from every controlled estimate.}\label{tab:pilot}
\begin{tabular}{lrrrr}\toprule
Family & Task & Episodes & Passed & Cost (\$) \\
\midrule
anthropic & gsm8k & 28 & 28 & 0.1841 \\
anthropic & minihouse & 28 & 28 & 0.3135 \\
deepseek & gsm8k & 28 & 28 & 0.0147 \\
deepseek & minihouse & 28 & 21 & 0.0127 \\
qwen & gsm8k & 28 & 28 & 0.0473 \\
qwen & minihouse & 28 & 23 & 0.0380 \\
\bottomrule\end{tabular}
\end{table}

\end{document}